\pdfoutput=1
\documentclass[sigconf,natbib]{acmart}

\AtBeginDocument{%
  }

\usepackage[utf8]{inputenc} 
\usepackage[T1]{fontenc}    
\usepackage{url}            
\usepackage{nicefrac}       
\usepackage{microtype}      
\usepackage{xcolor}         
\usepackage{bm}
\usepackage{bbm}
\usepackage{todonotes}
\usepackage{algorithm}
\usepackage{algpseudocode}
\usepackage{multicol}
\usepackage{caption}
\usepackage{multirow}
\usepackage{makecell}
\usepackage{booktabs}           
\usepackage{amsfonts}           
\usepackage{graphicx}           
\usepackage{duckuments}         
\usepackage{bold-extra}
\usepackage{amsmath}

\setcopyright{acmlicensed}
\copyrightyear{2025}
\acmYear{2025}
\acmDOI{10.1145/3746252.3760948}
\acmConference[CIKM '25] {Proceedings of the 34th ACM International Conference on Information and Knowledge Management}{ November 10--14, 2025}{Seoul, Republic of Korea.}
\acmBooktitle{Proceedings of the 34th ACM International Conference on Information and Knowledge Management (CIKM '25), November 10--14, 2025, Seoul, Republic of Korea}
\acmISBN{979-8-4007-2040-6/2025/11}
\acmDOI{10.1145/3746252.3760948}
\acmISBN{978-1-4503-XXXX-X/2018/06}

\acmSubmissionID{8928}

\begin{document}

\title{FUTURE: Flexible Unlearning for Tree Ensemble} 

\author{Ziheng Chen}
\email{albertchen1993pokemon@gmail.com}
\orcid{0000-0002-2585-637X}
\affiliation{%
  \institution{Walmart Global Tech, USA}
  \country{}
}

\author{Jin Huang}
\email{jin.huang@stonybrook.edu}
\affiliation{%
  \institution{Stony Brook University, USA}
  \country{}
}

\author{Jiali Cheng}
\email{jiali\_cheng@uml.edu}
\affiliation{%
  \institution{University of Massachusetts Lowell, MA, USA}
  \country{}
}

\author{Yuchan Guo}
\email{yuchanguo2001@gmail.com}
\affiliation{%
  \institution{Apple Inc, CA, USA}
  \country{}
}

\author{Mengjie Wang}
\email{mengjie225wang@gmail.com}
\affiliation{%
  \institution{Independent Researcher, NY, USA}
  \country{}
}

\author{Lalitesh Morishetti}
\email{lalitesh.morishetti@walmart.com}
\affiliation{%
  \institution{Independent Researcher, NY, USA}
  \country{}
}

\author{Kaushiki Nag}
\email{Kaushiki.Nag@walmart.com}
\affiliation{%
  \institution{Walmart Global Tech, USA}
  \country{}
}

\author{Hadi Amiri}
\email{Hadi_Amiri@uml.edu}
\affiliation{%
  \institution{University of Massachusetts Lowell, USA}
  \country{}
}

\renewcommand{\shortauthors}{Ziheng Chen, Fabrizio Silvestri, Jia Wang, Yongfeng Zhang, \& Gabriele Tolomei}

\newcommand{\Prob}{\mathbb{P}}
\newcommand{\Z}{\mathbb{Z}}
\newcommand{\insta}{\bm{x}}
\newcommand{\X}{X}
\newcommand{\G}{\mathcal{G}}
\newcommand{\advG}{\widetilde{\G}} 
\newcommand{\Gobs}{\G^{\text{obs}}}
\newcommand{\U}{\mathcal{U}}
\newcommand{\I}{\mathcal{I}}
\newcommand{\V}{\mathcal{V}}
\newcommand{\R}{\mathbb{R}}
\newcommand{\E}{\mathcal{E}}

\newcommand{\Vnew}{\V^{\text{new}}}
\newcommand{\Vadv}{\V^{\text{adv}}}
\newcommand{\edges}{\mathcal{E}}
\newcommand{\edgesnew}{\edges^{\text{new}}}
\newcommand{\edgesobs}{\edges^{\text{obs}}}
\newcommand{\edgesadv}{\edges^{\text{adv}}}
\newcommand{\graph}{\G=(\V,\edges)}
\newcommand{\bgraph}{\G=(\U, \I, \edges)}
\newcommand{\neigh}{\mathcal{N}}
\newcommand{\adjM}{A}
\newcommand{\advadjM}{\widetilde{A}}
\newcommand{\adjMij}{{A}_{i,j}}
\newcommand{\dataset}{\mathcal{D}}
\newcommand{\train}{\dataset_{\text{train}}}
\newcommand{\test}{\dataset_{\text{test}}}
\newcommand{\features}{\mathcal{X}}
\newcommand{\labels}{\mathcal{Y}}
\newcommand{\hypspace}{\mathcal{H}}
\newcommand{\params}{\bm{\theta}}
\newcommand{\w}{\bm{\omega}}
\newcommand{\h}{\bm{h}}
\newcommand{\advh}{\widetilde{\bm{h}}}
\newcommand{\hyp}{h_{\params}}
\newcommand{\gnn}{g(\adjM, \X; \W)}
\newcommand{\model}{h^*}
\newcommand{\loss}{\ell}
\newcommand{\Cons}{\mathcal{L}_{\text{cstr}}}
\newcommand{\ladv}{\ell_{\text{adv}}}
\newcommand{\ldist}{\ell_{\text{dist}}}
\newcommand{\lnew}{\ell_{\text{new}}}
\newcommand{\Loss}{\mathcal{L}}
\newcommand{\LF}{\Loss_{fa}}
\newcommand{\LC}{\Loss_{cf}}
\newcommand{\ind}{\mathbbm{1}}
\newcommand{\forget}{\mathcal{D}_f}
\newcommand{\remain}{\mathcal{D}_r}
\newcommand{\weights}{\boldsymbol{\omega}}
\newcommand{\ori}{g}
\newcommand{\neuori}{\tilde{g}}
\newcommand{\un}{g^{u}}
\newcommand{\optimal}{g^{*}}
\newcommand{\Tori}{\mathcal{T}}
\newcommand{\neuTori}{\tilde{\mathcal{T}}}
\newcommand{\Tun}{\mathcal{T}^{u}}
\newcommand{\Toptimal}{\mathcal{T}^{*}}
\newcommand{\inputs}{\mathcal{X}}
\newcommand{\outputs}{\mathcal{Y}}

\newcommand{\gabri}[1]{\todo[inline,color=red!60]{{\bf Gabri:} #1}}
\newcommand{\fabri}[1]{\todo[inline,color=green!60]{{\bf Fabri:} #1}}
\newcommand{\ziheng}[1]{\todo[inline,color=blue!60]{{\bf Ziheng:} #1}}
\newcommand{\Jiali}[1]{\todo[inline,color=pink!60]{{\bf Jiali:} #1}}
\newcommand{\La}[1]{\todo[inline,color=orange!60]{{\bf Lalitesh:} #1}}
\newcommand{\Hadi}[1]{\todo[inline,color=purple!60]{{\bf Hadi:} #1}}

\begin{abstract}
Tree ensembles are widely recognized for their effectiveness in classification tasks, achieving state-of-the-art performance across diverse domains, including bioinformatics, finance, and medical diagnosis. With increasing emphasis on data privacy and the \textit{right to be forgotten}, several unlearning algorithms have been proposed to enable tree ensembles to forget sensitive information. However, existing methods are often tailored to a particular model or rely on the discrete tree structure, making them difficult to generalize to complex ensembles and inefficient for large-scale datasets. 

To address these limitations, we propose FUTURE, a novel unlearning algorithm for tree ensembles. Specifically, we formulate the problem of forgetting samples as a gradient-based optimization task. In order to accommodate non-differentiability of tree ensembles, we adopt the probabilistic model approximations within the optimization framework. This enables end-to-end unlearning in an effective and efficient manner. Extensive experiments on real-world datasets show that FUTURE yields significant and successful unlearning performance.
\end{abstract}




\maketitle
\section{Introduction}
\label{sec:intro}

Advances in machine learning have significantly contributed to progress in a wide range of domains.
In particular, tree-based ensembles have emerged as a powerful approach for addressing classification tasks, especially in tabular data -- fundamental for applications in finance and healthcare. However, the widespread adoption of tree-based ensembles raises concerns about privacy leakage~\cite{zhang2025research,liu2024please,gao2020sensing}, as training data containing sensitive relationships can be implicitly ``memorized'' within model parameters. To mitigate the risk of misuse, recent regulatory policies have established the \textit{right to be forgotten}~\cite{chen2024debiasing,cheng2025multidelete}, allowing users to remove private data from online platforms. Consequently, a range of tree-based ensemble unlearning methods have been developed to effectively erase specific knowledge from a trained model without requiring full retraining~\cite{Liu2025}.

Machine unlearning \cite{bourtoule2021machine,Liu2025,fan2025towards,cheng25d_interspeech,cheng2025understanding,huang2025prompt,liu2025examining,cheng2025tool} for tree-based ensembles has recently gained considerable attention. Several works, including~\cite{brophy2021machine} and~\cite{schelter2021hedgecut}, focus specifically on unlearning in Random Forests(RF). ~\cite{brophy2021machine} selectively adjusts subtrees by modifying only the necessary upper-level split nodes. In contrast, ~\cite{schelter2021hedgecut} introduces a robustness quantification factor to identify splits that remain stable when a limited number of training instances are removed, enabling low-latency unlearning. ~\cite{lin2023machine} and ~\cite{wu2023deltaboost} are proposed for unlearning in Gradient Boosting Decision Trees (GBDT). Both algorithms exploit the internal tree structure and updates the threshold by incremental calculation in split gains and derivatives, resulting in a high computational cost for unlearning individual instances. In particular, ~\cite{lin2023machine} assumes that less than $1\%$ of the data will be removed, as the method becomes time-consuming for larger removal requests. In short, existing methods require full access to the models' training mechanisms and only work with specific tree-based ensembles. Moreover, their reliance on discrete structures hinders gradient-based updates and makes them scale poorly with the size of the forgetting set and the model.

In this work, we proposed FUTURE, an approximate unlearning algorithm designed to operate on any tree-based ensemble classifier. Inspired by~\cite{irsoy2012soft,dong2021gradient,liu2024please}, we first approximate the tree ensemble using a soft decision forest~\cite{kontschieder2015deep,lucic2022focus,luo2021sdtr}, replacing each split with a differentiable sigmoid function. This surrogate is trained end-to-end for the unlearning task, and the learned thresholds are ultimately transferred back to update the original ensemble.
Overall, the main contributions of our work are as follows:
\begin{itemize}
\item We unveil the limited generality and poor scalability of existing methods, and propose FUTURE, the first \textit{model-agnostic} unlearning algorithm designed for tree ensembles.

\item  FUTURE is an end-to-end unlearning framework for tree ensembles by leveraging a probabilistic approximation of their discrete structure, enabling scalability with respect to both ensemble size and forgetting set.

\item We conduct experiments on several real datasets, demonstrating that FUTURE can effectively and efficiently remove data as well as maintaining 95\% predictive power on test set, even when deleting 40\% of data.

\end{itemize}

\section{Background and Preliminaries}
\label{sec:background}
\subsection{Machine Unlearning}
In this work, we focus on machine unlearning for classification. This task involves selectively removing specific instances from a well-trained tree ensemble $\ori$ without retraining the entire model. We denote the set of samples to forget as $\mathcal{D}_f \subset \mathcal{D}$. 
The set of samples to be retained is denoted as $\remain$, with the assumption that $\mathcal{D}_f \cup \remain = \mathcal{D}$ and $\mathcal{D}_f\cap \mathcal{D}_r = \emptyset$~\cite{cheng2024mubench,chen2025frog,gao2024online}. The goal of machine unlearning is to obtain a new model $\un$ that removes the information contained in $\mathcal{D}_f$ from $\ori$ without adversely affecting its performance on $\remain$~\cite{zhang2021none}. Given the time-consuming nature of fully retraining the model on $\remain$ to obtain an retrained model $\optimal$, our target is to approximate $\optimal$ by evolving $\ori$ using $\mathcal{D}_f$ as follows:
\[
\ori \xrightarrow{\forget} \un \simeq \optimal
\]
\subsection{Tree-Based Ensembles}
\begin{figure}
    \centering
    \includegraphics[width=0.8\columnwidth]{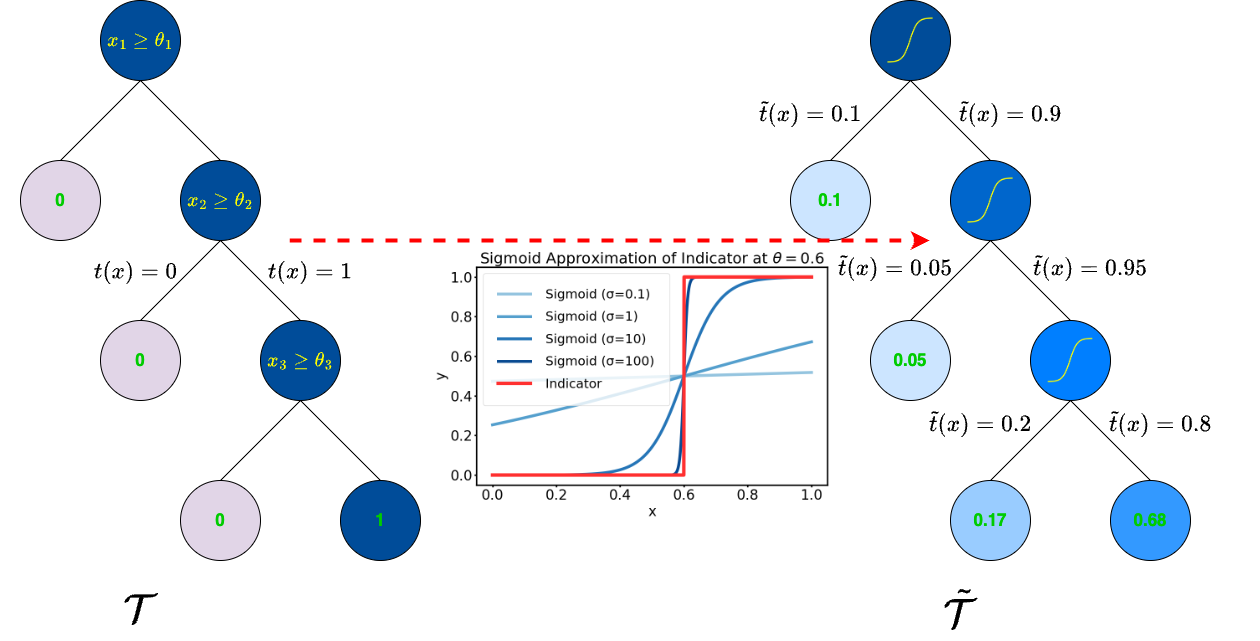}
    \caption{Approximating Tree-Based Ensembles With Soft Decision Forest. Left: Original Tree Ensembles $\mathcal{T}$. Right: Soft Decision Forest $\tilde{\mathcal{T}}$}
 \vspace{-0.5cm}   \label{fig:tree}
\end{figure}

  The tree ensemble $\ori$ can be viewed as a set of trees with weights $w_i$. For any sample $\mathbf{x}$, the tree ensemble return the weighted vote of its trees
\begin{equation}
\label{eq:ensemble}
\ori(y|\mathbf{x})=\arg\max\limits_{y}\sum^{M}_{i=1} w_i\cdot  \mathcal{T}_i(y|\mathbf{x}), 
\end{equation}
where $\mathcal{T}_i : \inputs \mapsto \outputs $ is a tree-structured classifier consisting of decision nodes and leaf nodes. As shown in figure ~\ref{fig:tree}, for one sample $\mathbf{x}$, a node $j$ is activated if its parent node $p_j$ is activated and feature $\mathbf{x}_{f_j}$ satisfies the threshold condition determined by $\theta_j$. Note that the root node is always activated. Let $t_{j}(\mathbf{x)}$ indicate if node $j$ is activated:

\begin{equation}
\label{eq:dt}
t_j(\mathbf{x}) = 
\begin{cases}
1, & \text{if } j \text{ is a root node} \\
t_{p_j}(\mathbf{x}) \cdot \mathbb{I}[\mathbf{x}_{f_j} \geq \theta_j], & \text{if } j \text{ is a right child of } p_j \\
t_{p_j}(\mathbf{x}) \cdot \mathbb{I}[\mathbf{x}_{f_j} < \theta_j], & \text{if } j \text{ is a left child of } p_j
\end{cases}
\end{equation}
 For an instance $\mathbf{x}$, it traverses from the root through activated decision nodes and ultimately reaches a single leaf node. Each leaf node $j$ is associated with a prediction distribution $t_l(y \mid j)$ that determines the output of $\mathcal{T}$ for $\mathbf{x}$.  Let $\mathcal{N}_l$ denotes the set of leaf nodes in $\mathcal{T}$, then we have:
 \begin{equation}
\label{eq:tree}
\mathcal{T}(y|\mathbf{x})=\sum_{j\in \mathcal{N}_l} t_j(\mathbf{x})\cdot t_{l}(y|j)
\end{equation}

In practice, all samples reaching the same leaf node are assigned the same class, regardless of their true distribution. Moreover, due to the discrete nature of the indicator function $\mathbb{I}$, $\Tori$ are not amenable to gradient-based training.

\section{Proposed Method: FUTURE}
\label{sec:method}

\begin{figure}
    \centering
 \includegraphics[width=0.95\linewidth]{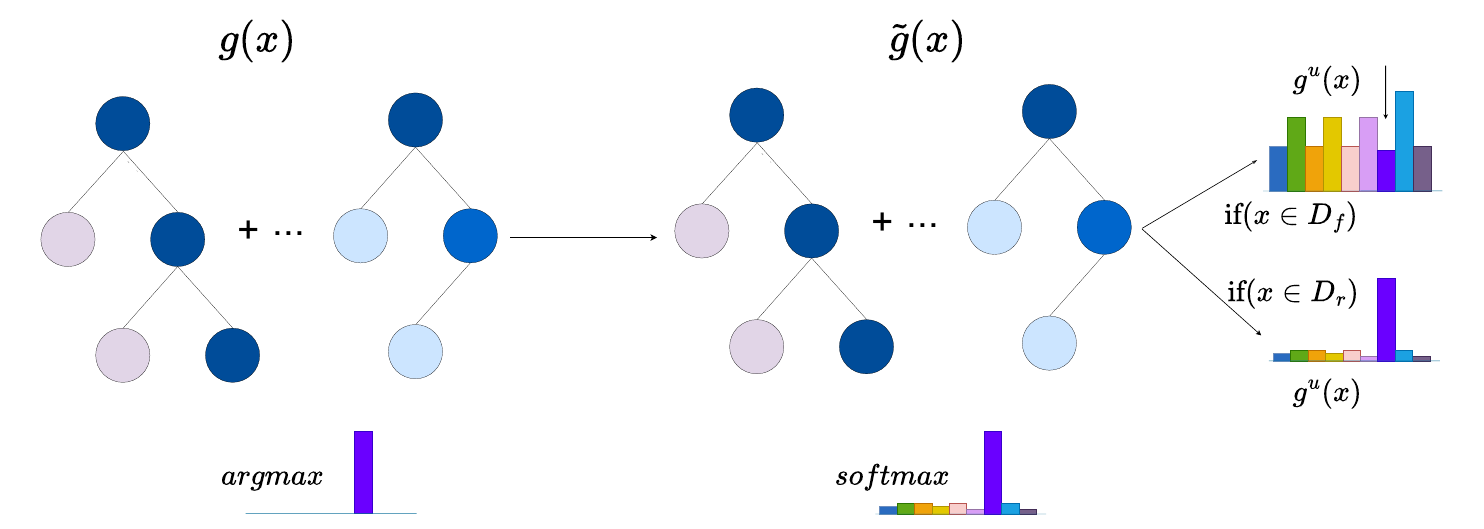}
    \caption{Our Proposed FUTURE unlearning algorithm}
 \vspace{-0.5cm}   \label{fig:FUTURE}
\end{figure}

This section presents FUTURE, a model-agnostic unlearning algorithm for tree ensemble $\ori$. As illustrated in Figure~\ref{fig:FUTURE}, we first construct a soft decision forest that emulates the $\ori$ without altering the structure of individual trees. We then formulate the unlearning task as a gradient-based optimization problem and update the thresholds of each tree accordingly. Notice that the soft decision forest serves purely as a surrogate to facilitate unlearning, rather than as a replacement for the original ensemble.



\begin{figure*}
    \centering
 \includegraphics[width=0.95\linewidth]{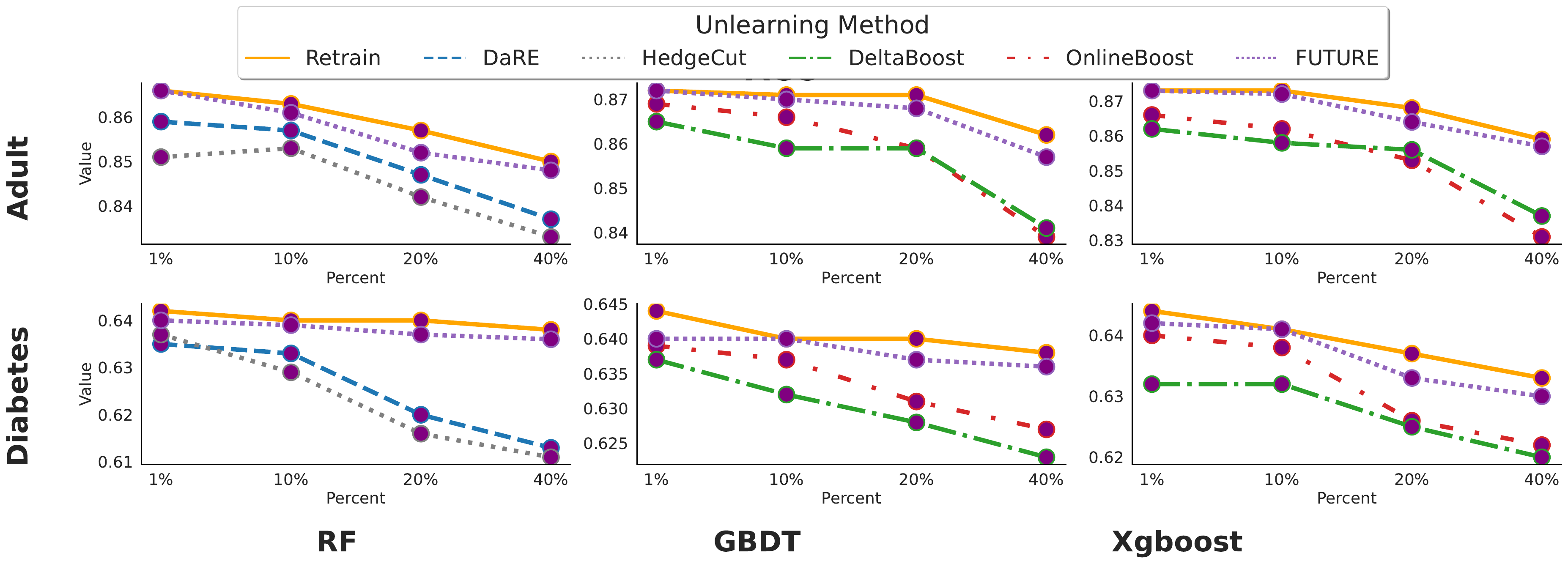}
    \caption{AOC-RUC on test set after unlearning}
 \vspace{-0.5cm}   \label{fig:acc}
\end{figure*}

\subsection{Approximating Tree-Based Ensembles with Soft Decision Forest}
\label{sec:appro}
To build a differentiable surrogate of $\ori$, it is crucial to capture the inherent logic of each decision tree it contains. To this end, we adopt the soft decision forest $\neuori$ as a surrogate for $\ori$.  $\neuori$ approximates the discrete operations in Equations~\ref{eq:ensemble} and~\ref{eq:dt}—specifically, the $\arg\max$ and the indicator function $\mathbb{I}$—using differentiable alternatives.

Inspired by~\cite{balestriero2017neural,lucic2022focus,lu2023machine,jayawardhana2025transformers}, the indicator function $\mathbb{I}[\mathbf{x}_{f_j} \geq \theta_j]$ can be approximated by the  sigmoid function $\text{sig}=(1+\text(exp)(-\sigma\cdot z))^{-1}$ with temperature parameter $\sigma$:$\mathbb{I}[\mathbf{x}_{f_j} \geq \theta_j]\approx \text{sig}(\mathbf{x}_{f_j}-\theta_j)$. As depicted in figure~\ref{fig:tree}, increasing $\sigma$ sharpens the sigmoid, making it a closer approximation to the indicator function. Based on this, we define the smoothed activation function $\tilde{t}_j$ by replacing all indicator functions accordingly:
\begin{equation}
\label{eq:ndt}
\tilde{t}_j(\mathbf{x}) = 
\begin{cases}
1, & \text{if } j \text{ is a root node} \\
\tilde{t}_{p_j}(\mathbf{x}) \cdot \text{sig}(\mathbf{x}_{f_j}-\theta_j), & \text{if } j \text{ is a right child of } p_j \\
\tilde{t}_{p_j}(\mathbf{x}) \cdot \text{sig}(\theta_j-\mathbf{x}_{f_j}), & \text{if } j \text{ is a left child of } p_j
\end{cases}
\end{equation}
Hence, we can introduce a neural decision tree approximation as follows:
\begin{equation}
\label{eq:neutree}
\tilde{\mathcal{T}}(y|\mathbf{x})=\sum_{j\in \mathcal{T}_l} \tilde{t}_j(\mathbf{x})\cdot t_{l}(y|j)
\end{equation}
Here $\mathcal{T}_l$ denotes the set of leaves. We retain the leaf prediction $t_l(y|j)$, as it follows a one-hot distribution. The surrogate model $\tilde{\mathcal{T}}$ preserves the tree structure and thresholds $\theta_j$ of the original tree $\mathcal{T}$. For an instance $\mathbf{x}$, the prediction $\tilde{\mathcal{T}}(y|\mathbf{x})$ in equation~\ref{eq:neutree} is no longer deterministic as in equation~\ref{eq:tree}, but represents the probability of reaching each leaf node $j$, depending on the distance between feature $\mathbf{x}_{f_j}$ and threshold $\theta_j$ at each decision node. 

Finally, we replace the maximum operation of ensemble $\ori$ by a softmax function with temperature $\tau$, resulting in:
\begin{equation}
\label{eq:neuensemble}
\tilde{\ori}(y|\mathbf{x})=\frac{\text{exp}\left(\tau\cdot\sum^{M}_{i=1} w_i\cdot  \tilde{\mathcal{T}}_i(y|\mathbf{x}) \right)}{\sum_{y^{'}}\text{exp}\left(\tau\cdot\sum^{M}_{i=1} w_i\cdot  \tilde{\mathcal{T}}_i(y^{'}|\mathbf{x}) \right)}
\end{equation}
This approximation could be generalized to any tree ensembles of its training strategy. The quality of the approximation relies on the choice of $\sigma$ and $\tau$. Although $\mathop{\lim}\limits_{\sigma, \tau \rightarrow \infty} \tilde{\ori}(y|\mathbf{x})=\ori(y|\mathbf{x})$, excessively large temperature parameters compromise smoothness and lead to vanishing gradients.  Therefore, we treat both $\sigma$ and $\tau$ as tunable hyperparameters.

\subsection{Unlearning on Neural Tree Ensembles}
Forgetting samples with tree structure is non-trivial because: i) the remaining samples $x_r\in\mathcal{D}_r$ and forgetting samples $x_f\in\mathcal{D}_f$ often traverse overlapping paths and share many decision and leaf nodes, making it difficult to isolate the influence of 
$x_f$ without affecting $x_r$ and ii) the final prediction of 
$\ori$ is determined by the weighted combination of all individual tree predictions. 

We begin by initializing $\un$ with a copy of $\tilde{g}$, which ensures good initial performance on both $\mathcal{D}_r$ and $\mathcal{D}_f$. Here, we treat all the thresholds in $\Theta=\{\theta \in \un \}$ as trainable parameters and freeze $\sigma$ and $\tau$.
To guarantee the effectiveness  of unlearning on the $\tilde{g}$, we follow two intuitive assumptions over the deleted data:

{\bf \emph{Deviation in Predictions over $\mathcal{D}_f$}}
To mitigate the influence of $x_f \in \mathcal{D}_f$, the prediction from the unlearned model $\un$ should deviate from that of $\tilde{\ori}$~\cite{zhang2024music}.Ideally, the prediction should resemble a uniform distribution over all labels, as if the model had never encountered samples from $\mathcal{D}_f$. As described in Section~\ref{sec:appro}, the probability of $x_f$ reaching each leaf node $j$ is given by the soft activation output $\tilde{t}_j$, computed by multiplying the soft activations along the path from the root. Hence we can derive the output distributions $p(\mathbf{x_f}, \tilde{g})$ and $p(\mathbf{x_f}, \un)$ via equation~\ref{eq:neuensemble}. To encourage this deviation, we maximize the predictive entropy on $\mathcal{D}_f$:
\begin{equation}
\label{eq:forgetloss}
\mathcal{H}_f(\mathbf{x_f},\un)=-\frac{1}{N_f}\sum_{\mathbf{x_f}\in \mathcal{D}_f} p(\mathbf{x_f},\un)\text{log}(p(\mathbf{x_f},\un))
\end{equation}
{\bf \emph{Stable Performance on $\mathcal{D}_r$}}
In practice, optimizing the forgetting loss inevitably degrades the performance of $\un$ on $\mathcal{D}_r$, as samples $x_r \in \mathcal{D}_r$ may also activate decision nodes affected by $\mathcal{L}_f$. To address this, we additionally encourage $\un$ to mimic the predictions of $\tilde{g}$ on $\mathcal{D}_r$ by minimizing the following loss:
\begin{equation}
\label{eq:remainloss}
\mathcal{L}_r(\mathbf{x_r},\un)=\frac{1}{N_r}\sum_{x_r\in \mathcal{D}_r} \text{KL}(p(\mathbf{x_r},\tilde{g})||p(\mathbf{x_r},\un))
\end{equation}
Here, KL denotes the KL divergence. To further strengthen the incentive to perform well, we also include the task loss $L_{cl}(\mathbf{x_r},\mathbf{y_r},\un)=l(p(\mathbf{x_r},\un),\mathbf{y_r})$ on $\mathcal{D}_r$, where $l$ stands for the cross-entropy loss and $\mathbf{y_r}$ is the true label of $\mathbf{x_r}$. Our training objective is then the following
\begin{equation}
\label{eq:forgetloss}\min\limits_{\un}\mathcal{L}_r(\mathbf{x_r},\un)+\alpha L_{cl}(\mathbf{x_r},\mathbf{y_r},\un)-\beta\mathcal{H}_f(\mathbf{x_f},\un)
\end{equation}
\subsection{Updating The Thresholds}
We update the thresholds $\Theta$ via batch learning. As previously described, $\un$ and $\ori$ share the same tree structure but differ in their thresholds. Thus, we can directly copy the trained $\Theta$ to $\ori$ to obtain the unlearned tree-based ensemble.

\section{Experiments}
\label{sec:experiments}

\subsection{Experimental Setup}
\noindent{\bf Datasets.}
In this paper, we conduct experiments following~\cite{lin2023machine} and ~\cite{brophy2021machine} over two datasets: \textit{Diabetes} and \textit{Adult}. \textit{Diabetes} contains 81412 instances with 43 attributes, and the objective is to predict whether or not a patient has diabetes or not. \textit{Adult} contains 48,842 instances with 14 attributes, and the objective is to predict if a user's income is above or below 50K. \\

\noindent{\bf Unlearning Setting.}
For each dataset, we randomly sample 1\%, 10\%, 20\%, and 40\% of data from $D$ as the forget set $D_f$. The rest is used as the retain set $D_r = D \setminus D_f$. \\

\noindent{\bf Unlearning Methods.}
We consider the following unlearning baselines. For random forests, we use DaRE~\cite{xin2019relational} and HedgeCut~\cite{xin2019relational}. For GBDT and XGBoost, we compare with DeltaBoost~\cite{wu2023deltaboost} and OnlineBoost~\cite{lin2023machine}. We also include retraining on $\mathcal{D}_r$ as a reference. \\

\noindent{\bf Metric.}
We assess unlearning quality using multiple metrics. To measure predictive performance, we report AUC-ROC ($\uparrow$) after unlearning \cite{cheng2023gnndelete}, which reflects any impact on model utility. Following~\cite{lin2023machine}, we evaluate unlearning effectiveness via backdoor attacks. Additionally, we measure unlearning efficiency by the time required for unlearning (Time, $\downarrow$).

\begin{figure}[ht]
\centering
\includegraphics[width=\columnwidth]{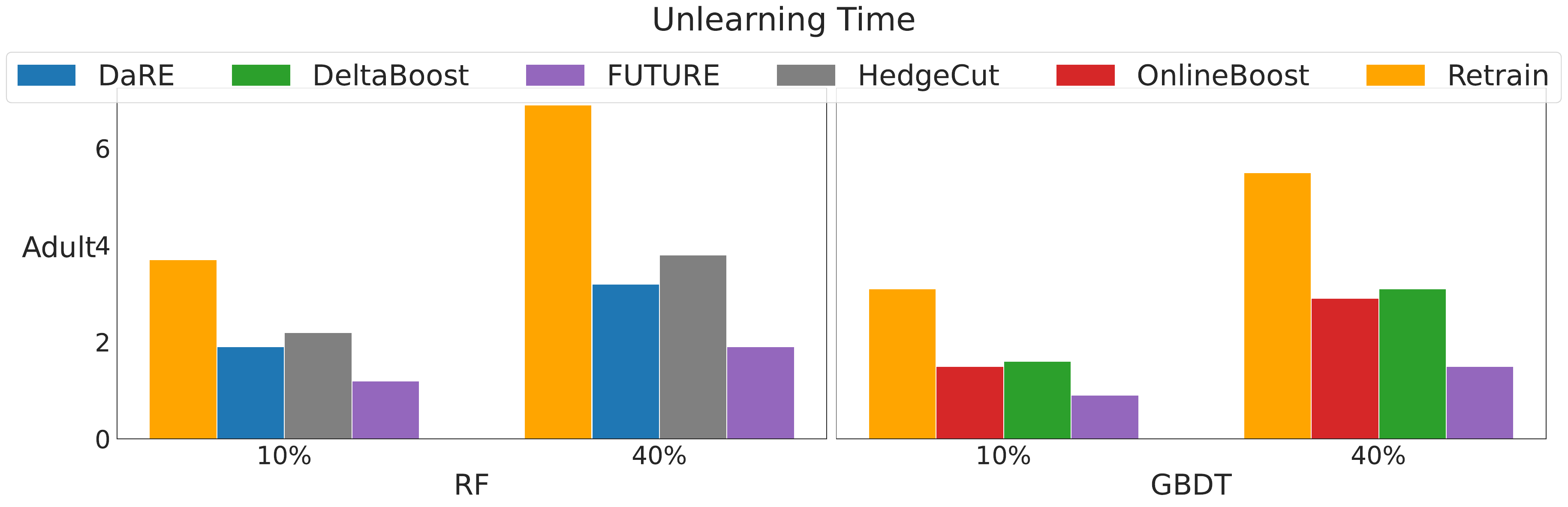}
\caption{Unlearning Time($10^{4}$ms), with $10\%$ and $40\%$ data removal}
\label{fig:Time}
 \vspace{-0.5cm} 
\end{figure}


\subsection{Main Results}

Overall, FUTURE preserving 95\% of AUC-ROC on test set, indicating it minimally affects model utility post-unlearning. Specifically, on average across all datasets and forget set site, FUTURE outperforms DaRE and HedgeCut by 0.015 and 0.021 on points on AUC-ROC Adult respectively, and by 0.024 and 0.026 points in AUC-ROC on Diabetes respectively. On GDBT, FUTURE outperforms DeltaBoost and OnlineBoost by 0.020 and 0.025 points on AUC-ROC on Adult respectively, and 0.014 and 0.020 points in AUC-ROC on Diabetes respectively. See Figure~\ref{fig:acc} for details. \\

\noindent{\bf FUTURE Maintains Unlearning Quality as Forget Set Increases}
When the forget set is small (1\%, 10\%), there is not a significant performance difference across different  unlearning methods. While the forget set increases to 20\% and 40\%, baselines has major performance degradation compared to Retrain. However, FUTURE can maintain 98\% of the test set performance. \\

\noindent{\bf FUTURE is Effective Across Multiple Tree-based Models} Following~\cite{lin2023machine}, we evaluate unlearning effectiveness through a backdoor attack. We randomly select $5\%$ of the training set, set a particular feature to the maximum value in the dataset, and assign label 1—treating these as poisoned samples. Models are first trained on the full dataset, and then various unlearning algorithms are applied to mitigate the poisoned influence. 

\begin{figure}[ht]
\centering
\includegraphics[width=\columnwidth]{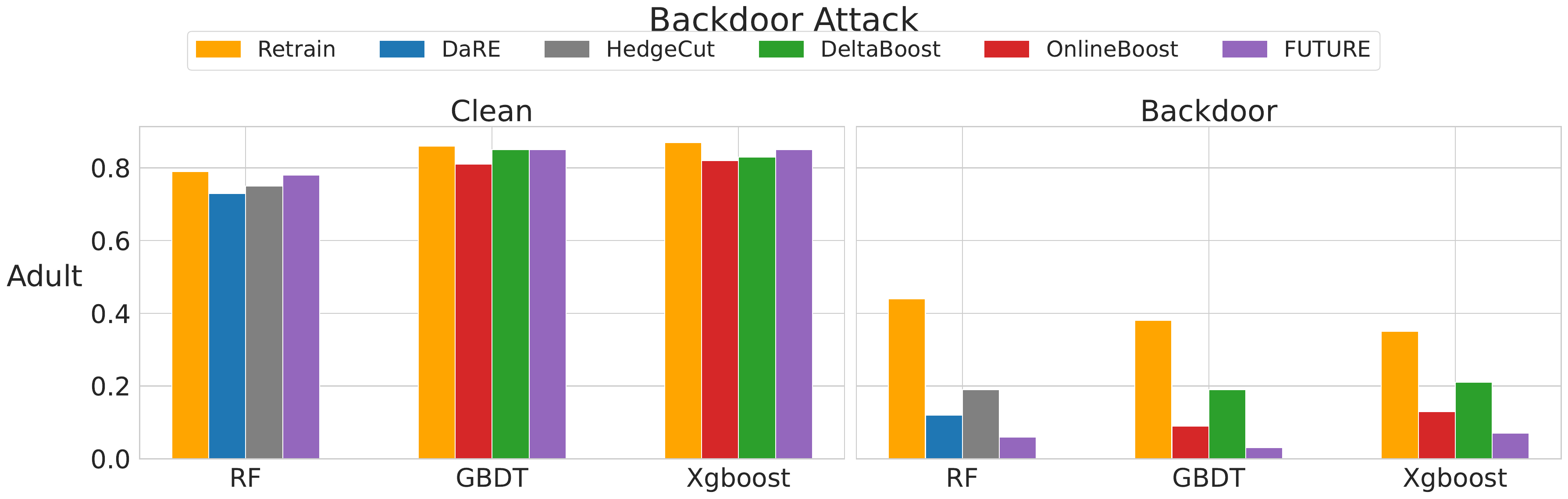}
\caption{Backdoor Removing Using Unlearning.}
\label{fig:backdoor}
 \vspace{-0.5cm} 
\end{figure}

Retraining can be seen as training on the clean subset. As shown in Figure~\ref{fig:backdoor}, the test accuracy on clean data remains comparable to training on clean data, outperforming OnlineBoost and DeltaBoost by $3\%$ and $2\%$, respectively. 

\noindent{\bf FUTURE is Efficient Across Multiple Tree-based Models} 
Besides the model performance, the cost of unlearning is also crucial for practical applications. As shown in figure~\ref{fig:Time}, we also compare our algorithm with others. When removing 10\% of data on RF, FUTURE saves 25, 07, and 10 seconds in training time compared to retraining from scratch, DaRE and HedgeCut. When removing 40\% of data, FUGURE saves 50, 13, and 19 seconds in training time compared to Retrain, DaRE and HedgeCut. 
When removing 10\% of data on GBDT, FUTURE saves 21, 6, and 7 seconds in training time compared to Retrain, DaRE and HedgeCut. When removing 40\% of data, FUGURE saves 40, 14, and 16 seconds in training time compared to Retrain, DaRE and HedgeCut. \\

\noindent{\bf FUTURE is a Model-Agnostic Unlearning Method}
As shown in the experiment, FUTURE can be applied to all kinds of tree-based ensembles including but not limited to RF, GBDT, Xgboost, which no baseline is able to achieve. DaRE and HedgeCut can work on RF but not GBDT or Xgboost, while DeltaBoost and OnlineBoost can work on boosting trees but not RF.

\section{Conclusion and Future Work}
\label{sec:conclusion}
We present FUTURE, a novel model-agnostic, effective and efficient tree-based unlearning methods. It can work with all kinds of tree-based models.
Compared to tree-based unlearning methods, it can be trained from start to finish, directly targeting unlearning requests and significantly improving unlearning quality.

\begin{acks}
This research was, in part, funded by the National Institutes of Health (NIH) Agreement No. 1OT2OD032581. The views and conclusions contained in this document are those of the authors and should not be interpreted as representing the official policies, either expressed or implied, of the NIH.
\end{acks}

\section*{GenAI Disclosure Statement}
We used GPT-4 to identify and correct grammatical errors, typos, and to improve the overall writing quality. 
No AI tools were used at any other stage of this work to ensure full academic integrity.

\bibliographystyle{ACM-Reference-Format}
\balance
\bibliography{references}










\end{document}